# Virtual-Blind-Road Following Based Wearable Navigation Device for Blind People

Jinqiang Bai, Shiguo Lian, *Member, IEEE*, Zhaoxiang Liu, Kai Wang, Dijun Liu

*Abstract*—To help the blind people walk to the destination efficiently and safely in indoor environment, a novel wearable navigation device is presented in this paper. The locating, way-finding, route following and obstacle avoiding modules are the essential components in a navigation system, while it remains a challenging task to consider obstacle avoiding during route following, as the indoor environment is complex, changeable and possibly with dynamic objects. To address this issue, we propose a novel scheme which utilizes a dynamic sub-goal selecting strategy to guide the users to the destination and help them bypass obstacles at the same time. This scheme serves as the key component of a complete navigation system deployed on a pair of wearable optical see-through glasses for the ease of use of blind people's daily walks. The proposed navigation device has been tested on a collection of individuals and proved to be effective on indoor navigation tasks. The sensors embedded are of low cost, small volume and easy integration, making it possible for the glasses to be widely used as a wearable consumer device.

*Index Terms*—Virtual blind road, obstacle avoiding, route following, wearable navigation device

## I. INTRODUCTION

THE visually impaired people usually have difficulties in walking in an unfamiliar and complex place independently. To provide them an automatic navigation device with effective guidance on their move, three problems should be considered:

1.) Where is the person? The device has to know where the person is located in order to make a correct decision for guiding the person. This refers to be the localization problem.

2.) Where does the person want to go? In order to help the visually impaired person reach his destination, the device has to identify the destination. This is known as goal recognition.

3.) How does the person get there? This includes way-finding, route following and obstacle detecting. Way finding is to plan a shortest path from the starting position to the destination, route following is to make sure the blind person follow the planned path and obstacle detecting is to help him avoid obstacles.

Manuscript received January 1, 2018; revised February 8, 2018; accepted February 15, 2018.
This work was supported by the CloudMinds Technologies Inc.
Jinqiang Bai is with Beihang University, Beijing, 10083, China (e-mail: baijinqiang@buaa.edu.cn).
Shiguo Lian, Zhaoxiang Liu, Kai Wang are all with AI Department, CloudMinds Technologies Inc., Beijing, 100102, China (e-mail: { scott.lian, robin.liu, kai.wang }@cloudminds.com).
Dijun Liu is with CATT, Beijing, 10083, China (e-mail: liudijun@datang.com).

So far, there are many navigation systems trying to solve the above problems, such as the low-cost white cane [1], guide dog [2] and ETAs (Electronic Travel Aids) [3]. However, white cane is unable to find a globally shortest path [2] and provide the location information. Guide dog is incapable of detecting overhanging object, and needs costly training, which may be unaffordable to the visually impaired individuals [1], [2]. Most existing ETAs are only intended for obstacle detecting or/and feedback [4-9], and cannot provide way-finding and route following functions. Although some ETAs were designed with way-finding and route following functions, such as the cactus tree based algorithm proposed in [10], obstacle (especially the dynamic obstacles) detecting and avoiding are ignored. A wearable indoor navigation system based on visual marker recognition and ultrasonic obstacle perception was introduced in [11], but the localization precision is not high enough for guiding the blind due to the error increase of the inertial measurement sensor, and the goal recognition scheme is less efficient on planning a global path. A successful navigation system for the blind is the visual SLAM (Simultaneous Localization and Mapping) and PoI (Point of Interest)-graph based indoor navigation system presented in [1], [12]. However, the obstacle detection heavily relies on the white cane swaying, which is not efficient and portable.

This paper aims to develop an effective wearable navigation system which can locate the user, follow the virtual-blind-road and avoid obstacles at the same time, in order to provide automatic navigation for the visually impaired people. The main contribution of this paper is the proposal of a novel dynamic sub-goal selecting based virtual-blind-road following scheme which combines the obstacle avoiding algorithm [13] for guiding the blind people to follow the globally shortest virtual-blind-road without collision. This scheme can help the visually impaired people to avoid obstacles while following a path precisely to the destination, which realizes the automatic navigation for them. As the GPS-based blind navigation technology [14] cannot be used due to the severely degradation of GPS signal in indoor environment, the visual SLAM (e.g. ORB-SLAM (ORiented Brief-SLAM) [15], [16]) was adopted in this paper for the building of the virtual-blind-road and the localization. The whole system is deployed on a pair of wearable optical see-through glasses, which is able to give the visually impaired people visual hints and sound feedback.

The rest of the paper is organized as follows. Section II reviews the related works on the navigation for the visually impaired in indoor environment. The hardware configuration of the proposed navigation device is presented in Section III. Section IV describes the navigation scheme in detail. Section V shows some experimental results, and demonstrates the effectiveness of the proposed blind navigation system. Finally, some conclusions are drawn in Section VI.

## II. RELATED WORK

Navigation for the blind in indoor environment should consider the problems of localization, way-finding, route following and obstacle detecting. Many ETAs were reviewed in details [13], which utilize ultrasonic, laser scan, camera or multi-sensor fusion technology for obstacle detecting and/or avoiding. These ETAs can help the visually impaired avoid the obstacles, but they are unable to provide the location information, and do not have way-finding and route following functions. Thus, the related works on localization, way-finding and route following for blind navigation in indoor environment are reviewed in this section.

### A. Indoor Localization

Indoor localization can be broadly classified into two categories: wireless localization and vision-based localization.

There are many wireless localization systems that utilize UWB (Ultra Wide Band) radar [17], RFID (Radio Frequency Identification) [18], ZigBee [19], Bluetooth [20] and Wi-Fi [21] [22]. UWB radar has been a particularly attractive technology for indoor localization because it is not susceptible to interference from other signals (due to its drastically different signal type and radio spectrum), and the UWB signal (especially the low frequencies included in the broad range of the UWB spectrum) can penetrate a variety of materials, including walls. UWB radar achieves very high localization accuracy, which can be up to 10cm. However, the slow progress in the UWB standard development has limited the use of UWB in consumer products and portable user devices. RFID based systems can be used for localization and object tracking as they have a reasonable range, low cost and can be easily embedded in the tracking objects. However, the limited range makes them unsuitable for high precision indoor localization. ZigBee is concerned with the physical and MAC (Media Access Control) layers for low cost, low data rate and energy efficient personal area networks. It is favorable for localization of sensors in WSN (Wireless Sensor Network), but it is not suitable for indoor localization as it cannot provide the distance between the user and the labeled sensor. Most of the existing Bluetooth based localization systems rely on RSSI (Received Signal Strength Indicator), however, they are prone to multipath fading and environmental noise, lower localization accuracy due to its signal band (2.4 G) which is the same with other signals. Thus, only Bluetooth based localization system has very large uncertainty in localization Most of portable consumer devices (e.g. smart phones, laptops) are Wi-Fi enabled, which makes Wi-Fi an ideal candidate for indoor localization. However, existing Wi-Fi networks are normally deployed for communication (i.e., to maximize data throughput and network coverage) rather than localization purposes. Therefore their localization accuracy still needs improvement. Moreover, all these systems require special hardware.

The vision-based localization takes advantage of image processing and objects recognition. An effective way for indoor navigation using IMU (Inertial Measurement Unit) in conjunction with QR (Quick Response) code was proposed in [23]. QR code based system requires proximity recognition [24], nevertheless, the blind people have difficulty in approaching the QR code. Therefore, it is inappropriate to blind localization. Recently visual SLAM technology including state filter based SLAM [25] and pose-graph SLAM [1], [12], [15], [16], [25-29] become more and more popular in indoor navigation as its localization error can be effectively reduced at a loop-closing point and the positioning accuracy can be up to the centimeter scale. Besides, the visual SLAM based localization systems require no other special hardware except for a camera, stereo camera or RGB-D camera. Therefore, the visual SLAM was utilized to locate the visually impaired individuals accurately in this paper.

### B. Way-Finding and Route Following

The sparse map built by visual SLAM module becomes a kind of PoI-graph [1] [12] after some PoIs were tagged on it. This PoI-graph was utilized in this paper as the global map. There are many PoI-graph based way-finding algorithms, such as Dijkstra [30], A* [1] [12], D* [31], etc. However, this is not the focus of this paper. It is just for providing the global path (i.e. the path to be followed. Hence, any of these algorithms can be used for planning the globally shortest path. In this paper, A* algorithm was adopted as the way-finding algorithm.

After the globally shortest path from the starting position to the destination was found, the route following algorithm will provide accurate guidance information to ensure that the user can follow the globally shortest path and avoid obstacles. The route following algorithm in [10] takes the globally shortest path and a priori knowledge of indoor environment as inputs, and outputs a control strategy that makes the user's trajectory remain close to the globally shortest path. It selects some special nodes (e.g. entrance node, room node and intersection node, etc.) as sub-goals to guide the user. The method [32] proposed a path deviation detecting method which is based on measuring the similarity between the current frame and the training frame of the pre-defined path. However, the dynamic obstacles and changes of the environment were not considered. Hence, it is not able to help the blind avoid dynamic obstacles. To overcoming the above problems, a dynamic sub-goal based route following algorithm was proposed in this paper. The sub-goal is selected dynamically, which includes but is not limited to the special node as mentioned in [10]. The above route following algorithm combining the obstacle detecting method [13] which takes the dynamic obstacles and changes of environment into account, and the globally shortest path, output the optimal walkable direction to ensure the user follow the globally shortest path closely, meanwhile, avoid obstacles.

## III. HARDWARE CONFIGURATION

The proposed navigation device consists of a fisheye and a depth camera, an ultrasonic rangefinder, an embedded CPU

board, a pair of OST (Optical See-Through) glasses, and an earphone. These sensors are of low cost, small volume, and easy integration, and thus can be widely used in consumer market.

The depth camera and the fisheye camera are used for building the virtual-blind-road and locating the user precisely with visual SLAM algorithm. Besides, the depth camera is also used for obstacle detection [13]. To compensate for the limitations of the depth camera, such as passing through the transparent objects, being absorbed by some special materials, etc., the ultrasonic rangefinder is utilized. It consists of an ultrasonic sensor and a MCU (Microprogrammed Control Unit) to measure the distance of an obstacle in front of the user. Fusing the depth camera and the ultrasonic rangefinder for obstacle detection can ensure adequate safety of the visually impaired individual. The CPU operates at 1.44 GHz with a 2 GB RAM. All the algorithms such as visual SLAM, obstacle detecting, way-finding, route following, speech synthesis, etc. are performed on the CPU. The OST-glasses have two display screens for displaying the guiding information and surrounding information to the partially sighted individual. In addition, the glasses have two loudspeakers for playing the guiding sound. The earphone is used for playing the guiding sound to the totally blind individuals when they are in a noisy indoor environment (e.g. supermarket). The hardware configuration of the proposed navigation device is illustrated in Fig. 1, and its prototype is shown in Fig. 2.

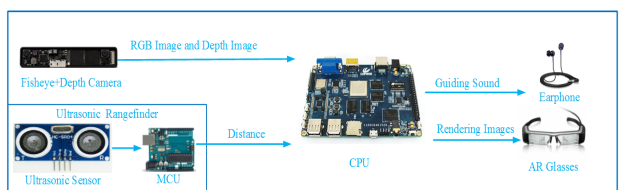

Fig. 1. The hardware configuration of the proposed navigation device.

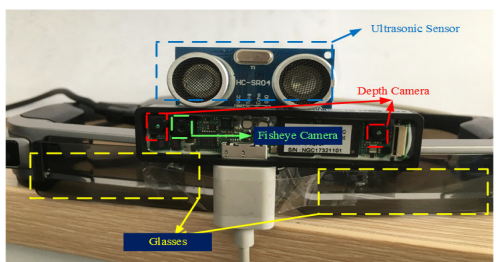

Fig. 2. The prototype of the proposed navigation device.

## IV. THE NAVIGATION SCHEME

The architecture of the proposed navigation system is illustrated in Fig. 3, which includes visual SLAM, PoI-graph, way finding, obstacle detection and route following module. The rendering module converts the guiding information produced by the navigation system into audio or/and visual cues. The obstacle detection and rendering modules have been presented in our previous work [13], and thus will not be repeated here.

The visual SLAM module utilizes the RGB image and depth image to build the virtual-blind-road (offline) and locate the user (online). After the virtual-blind-road is tagged with some key positions (offline), it becomes a sparse map (i.e. PoI-graph). The way finding module plans a globally shortest path from the starting position to the destination according to the PoI-graph. The obstacle detection module provides many candidate walkable directions according to the current depth image. The route following module takes the current pose of the blind user, the globally shortest path, the candidate walkable directions, and the obstacle distance measured by the ultrasonic rangefinder as inputs, and uses a dynamic sub-goal selecting method to produce the guiding information. This allows the user to follow the globally shortest path as closely as possible and avoid obstacles at the same time.

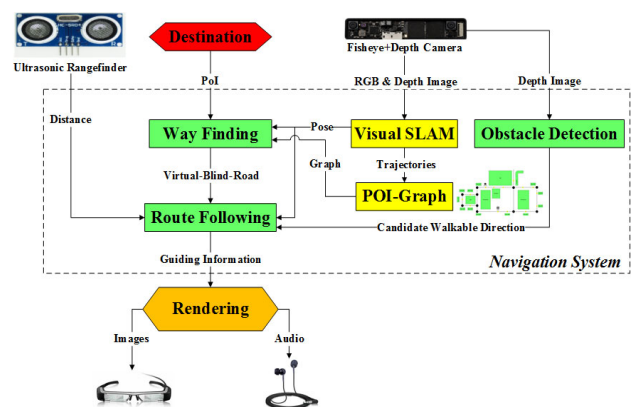

Fig. 3. Architecture of the proposed navigation system.

### A. Visual SALM and PoI-graph

The ORB-SLAM2 [16] was utilized in this paper. It has three main parallel threads: Tracking, Local Mapping and Loop Closing. The tracking thread can locate the camera through feature matching with the local map. The local mapping thread can insert new point into the local map and optimize the local map by performing local BA (Bundle Adjustment). The loop closing thread can detect large loops, and optimize the pose graph. Besides, the loop closing thread will launch a fourth thread to perform full BA after the pose graph optimization, to compute the optimal global structure and motion solution.

In this paper, the ORB-SLAM2 was used to solve two main problems:

1.) Building the virtual-blind-road. Before providing navigation information for the visually impaired, it is necessary to build the virtual-blind-road. The virtual-blind-road is built by a sighted person wearing the proposed navigation device and performing all threads of the ORB-SLAM2. The virtual-blind-road (as is shown in Fig. 4 (a)) is actually the motion trajectory of the sighted person. The process of building the virtual-blind-road is completed offline.

2.) Locating the user. When navigating the visually impaired to walk from the starting position to the destination, only the tracking thread of the ORB-SLAM2 is operated for localization. The 3D pose of the camera can be computed through feature matching between the current frame and previous frame or key-frame in the pre-built map [16]. The matches between current frame and previous frame can cope with the unmapped regions in the pre-built map, however, this will cause the accumulation of drift. Nevertheless, once the current frame matched with the key-frame in the pre-built map, the drift will be corrected. Thus, the localization can be robust and accurate. Besides, the fisheye camera can also improve the locating accuracy by providing more feature matching points due to its wide angle of view.

After the virtual-blind-road is built, the PoI-graph will be generated through tagging some PoI such as rooms, toilet, bar, hallway junctions as the nodes on the virtual-blind-road. The edge connecting two nodes in PoI-graph has a weight equal to the real distance between the nodes. Taking an office area as an example, the PoI-graph is shown in Fig. 4 (b). Each edge corresponds to a piece of virtual-blind-road and each node represents a place of interest. The PoI-graph is used for finding a globally shortest path from the starting position to the destination (see section IV. B).

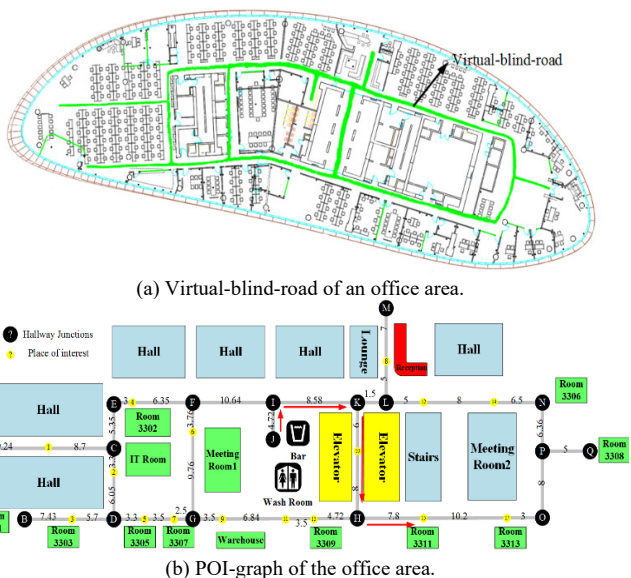

(a) Virtual-blind-road of an office area.

(b) POI-graph of the office area.

Fig. 4. Example of virtual-blind-road and corresponding POI-graph.

### B. Way Finding

The way finding module plans a globally shortest path by applying A* algorithm to the PoI-graph. A* algorithm searches among all possible paths to the destination for the one that incurs the smallest cost (shortest distance). For instance, the globally shortest path from the bar to Room 3311 is depicted by the red arrows in Fig. 4(b) (J->I->K->H->Room 3311).

### C. Route Following

The route following module can be described as:
Inputs:
1.) A globally shortest path (i.e. the virtual-blind-road) produced by way finding module.
2.) The current pose output by the visual SLAM module.
3.) Candidate walkable directions generated from the obstacle detection module.
4.) Obstacle distance measured by the ultrasonic rangefinder.

Output: Guiding information that ensures the user follow the globally shortest path as closely as possible, and avoid obstacles at the same time.

Terms:
1.) Global path: $W = \{w_1(x_1, y_1), w_2(x_2, y_2), \ldots, w_n(x_n, y_n)\}$, a sequence of position, which is capable of guiding the visually impaired individual from the current position to the destination, where $w_i(x_i, y_i)$ is a point defined in the global coordinate system.
2.) Current pose: $P_c(x_c, y_c, \theta_c)$, current position and orientation, where $(x_c, y_c)$ is current coordinate in global coordinate system and $\theta_c$ is current heading (i.e. the direction of camera optical axis).
3.) Sub-goal: the navigation point $P_G(x_G, y_G)$. This is the desired position next step.
4.) Candidate walkable directions: the walkable directions $D = \{\theta_1, \theta_2, \ldots, \theta_n\}$ produced from the obstacle detection, $\theta_i$ is the angle relative to the current heading (i.e. the direction of camera optical axis).
5.) Obstacle distance: the distance $d$ measured by the ultrasonic rangefinder. If an obstacle is in front of the user and in the view field of the ultrasonic rangefinder, the distance will be used for guiding the user safely.
6.) Guiding direction: the optimal walkable direction $\theta_{opt}$. It is relative to the current heading (i.e. the direction of camera optical axis), then will be converted to audio and/or visual cue for guiding the visually impaired individual.

Based on the above descriptions, the steps of the proposed route following algorithm are as follows.

Step 1: get the current pose $P_c$ and the globally shortest path $W$.

Step 2: select the sub-goal $P_G$ according to the current pose and the globally shortest path. The sub-goal is selected as follows:

Firstly, find the closest point to the current position in the globally shortest path, which can be formulized as:

$$P_{cls}(x, y) = \arg\min_{(x_i, y_i) \in W} \sqrt{(x_i - x_c)^2 + (y_i - y_c)^2}, \quad (1)$$

where $P_{cls}(x, y)$ is closest point to the current position in the globally shortest path, $W$ is the globally shortest path, $(x_c, y_c)$ is the current position.

Secondly, select the sub-goal, which can be a point having a certain distance from the closest point, a turning point or destination point, this can be calculated as:

$$P_G(x_G, y_G) = \{(x_s, y_s) \mid \sum_{i=1}^{s}\sqrt{(x_i - x_{cls})^2 + (y_i - y_{cls})^2} \geq d \text{ P}$$

$$\arccos\frac{(x_s - x_{cls})(x_{s+1} - x_s) + (y_s - y_{cls})(y_{s+1} - y_s)}{\sqrt{(x_s - x_{cls})^2 + (y_s - y_{cls})^2}\sqrt{(x_{s+1} - x_s)^2 + (y_{s+1} - y_s)^2}} \geq \theta \parallel \quad (2)$$

$(x_{last}, y_{last}), (x_i, y_i) \in W, i > cls\}$,

where $P_G(x_G, y_G)$ is the selected sub-goal (the blue dot in Fig. 5), $d$ is the distance threshold, $W$ is the globally shortest path, $(x_{cls}, y_{cls})$ is the closest point (the yellow dot in Fig. 5) to the current position (the red dot in Fig. 5), $\theta$ is the angle threshold which can be considered as a turning point, $(x_{last}, y_{last})$ is the last point of the globally shortest path $W$, i.e. the destination point. The cases which a point can be selected as the sub-goal are shown in Fig. 5:

Case 1: A point $P_G$ is collinear or approximately collinear with the closest point $P_{cls}$, select the point $P_G$ that its distance to the closest point exceeds the threshold $d$ (e.g. 3 m) as the sub-goal. Comparing with key node-based navigation method, this ensures the users follow the path closely, and the users can be informed when they deviate from the global path.

Case 2: A point $P_G$ that its distance to the closest point $P_{cls}$ does not exceed the threshold $d$, but the angle between $P_{cls}P_G$ and $P_G P_{G+1}$ exceeds the threshold $\theta$ (e.g. $\pi/6$). This point is considered as a turning point, which can be taken as the sub-goal. This kind of sub-goal does not update until the user reaches this point. This ensures the user safely get around a corner.

Case 3: This point is the final point of the globally shortest path in section IV. B, which does not satisfy the above two cases. It will be selected as the sub-goal before it was reached. If the distance between the current position and this point is less than a small threshold, such as 1.0 m, it is considered as reaching the destination.

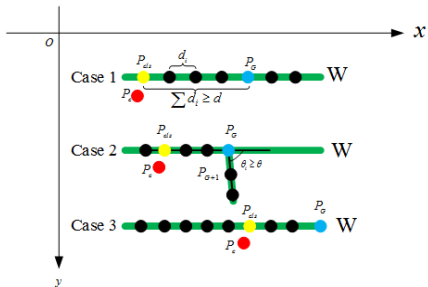

Fig. 5. Cases of sub-goal selecting.

Step 3: compute the candidate walkable directions $D=\{\theta_1, \theta_2, L, \theta_n\}$ according to current depth image [13].

Step 4: compute the final walkable direction. Firstly, according to the current pose and the sub-goal, the expected direction or angle can be calculated as:

$$\theta_{exp} = \begin{cases} \dfrac{\pi}{2} & if(x_G - x_c) = 0, (y_G - y_c) > 0 \\ \dfrac{3\pi}{2} & if(x_G - x_c) = 0, (y_G - y_c) < 0 \\ \arctan\dfrac{(y_G - y_c)}{(x_G - x_c)} & if(x_G - x_c) > 0, (y_G - y_c) > 0 \\ \arctan\dfrac{(y_G - y_c)}{(x_G - x_c)} + 2\pi & if(x_G - x_c) > 0, (y_G - y_c) < 0 \\ \arctan\dfrac{(y_G - y_c)}{(x_G - x_c)} + \pi & else \end{cases} \quad (3)$$

where $\theta_{exp}$ is the expected angle, $(x_c, y_c)$ is the coordinate of current position, $(x_G, y_G)$ is coordinate of sub-goal. The angle value is in $[0, 2\pi)$, which is depicted in Fig. 6.

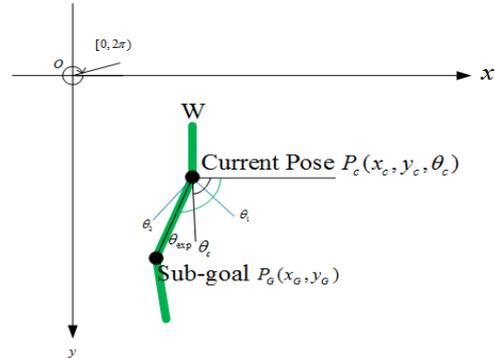

Fig. 6. Expected angle representation.

Then, based on the candidate walkable directions $D=\{\theta_1, \theta_2, L, \theta_n\}$, expected direction $\theta_{exp}$ and current pose $P_c(x_c, y_c, \theta_c)$, the optimal walkable direction can be obtained by minimizing the cost function, which is defined as:

$$\theta_{opt} = \begin{cases} Null & if(D=\varnothing) \\ \underset{\theta_i \in D}{\arg\min} \|\theta_{exp} - \theta_c - \theta_i\| & if(D \neq \varnothing) \&\& \\ & (\underset{(x_i, y_i) \in W}{\min} \sqrt{(x_i - x_c)^2 + (y_i - y_c)^2} > 1.0\,P, \\ & \|\theta_{exp} - \theta_c\| \geq \dfrac{\pi}{6}) \\ \underset{\theta_i \in D}{\arg\min} \|\theta_i\| & else \end{cases} \quad (4)$$

where $\theta_{opt}$ is the optimal walkable direction, $W$ is the globally shortest path. The intuition of Eq. 4 is that when the visually impaired deviated from the globally shortest path, i.e. the minimum distance between the current position and the globally shortest path exceeds a threshold (e.g. 1.0 m in this work), or the angle between the current heading and the expected direction exceeds a threshold (e.g. $\pi/6$ in this work), the way following module will correct the user toward the globally shortest path; otherwise, the optimal direction will be selected only from the candidate walkable directions.

Next, ultrasonic rangefinder is fused to determine the final walking direction in order to compensate for the limitations of the depth camera [13]. Since the ultrasonic sensor can detect the obstacles in the range of 0.03 m to 4.25 m, and within scanning field of 15°, the final walking direction is defined as:

$$\theta_{walk} = \begin{cases} \theta_{opt} & if(\theta_{opt} \notin [-7.5, 7.5])\,P(\theta_{opt} \in [-7.5, 7.5]\,\&\&\,d_{ultra} > \delta) \\ Null & else \end{cases} \quad (5)$$

where $\theta_{walk}$ is the final walking direction, $d_{ultra}$ is the distance measured by the ultrasonic rangefinder, $\delta$ is the distance threshold of obstacle (e.g. 2.0 m in this work). It judges if the optimal walkable direction $\theta_{opt}$ is within the view field of ultrasonic sensor, i.e. [-7.5°, 7.5°]. If not, it will directly output the optimal walkable direction, otherwise, the ultrasonic rangefinder then will be used to judge if the measured distance exceeds the threshold $\delta$. If yes, it will also output the optimal walkable direction; conversely, it will output Null, which means no walkable direction in the field of view. This guiding direction will then be converted to audio and/or visual cue for guiding the visually impaired individual [13].

These four steps are continually executed until the destination is reached. The workflow of the proposed navigation system is shown in Fig. 7.

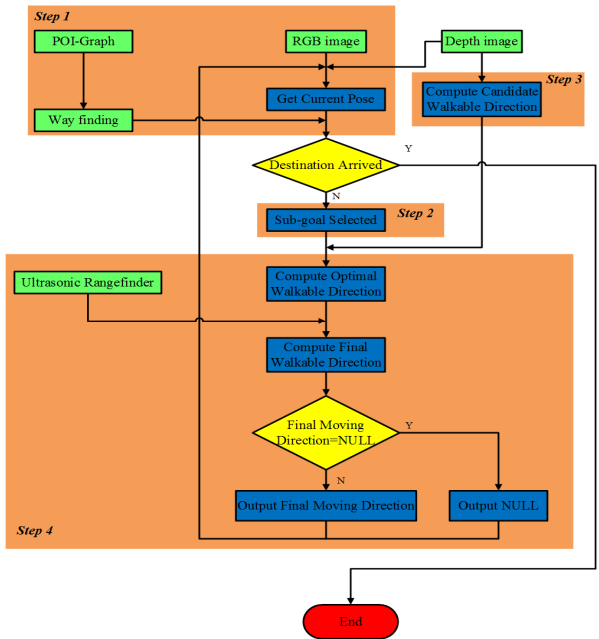

Fig. 7. The workflow of the proposed navigation system.

## V. EXPERIMENTAL RESULTS AND DISCUSSIONS

To test the effectiveness of the proposed navigation system, two groups of experiments were conducted, including Experiment 1 without obstacles and Experiment 2 with obstacles. In both experiments, 3 paths (from A (hall) to B (wash room), D (lounge) to C (bar) and E (Room 3308) to B), as shown in Fig. 8 and Fig. 9 are provided. Individuals of totally blind and partially sighted were navigated by the proposed system to follow these paths, and their actual walking trajectories were recorded to analyze both qualitatively and quantitatively.

Experiment 1: there are no obstacles on the 3 paths. The users can follow the path to destination, as shown in Fig. 8. For quantitative analysis of the route following algorithm, the corresponding maximum, the average distance deviating from the 3 paths and the variance were calculated, and the results are shown in Table I. The maximum deviation distance is less than 1.0 m and the average deviation distance is less than 0.3 m, which demonstrates that the proposed algorithm can navigate the visually impaired effectively by following the virtual-blind-road. Besides, the variance is very small, which additionally verified the effectiveness of the route following algorithm. Once the user deviated from the globally shortest virtual-blind-road, the route following module will inform the user to get back.

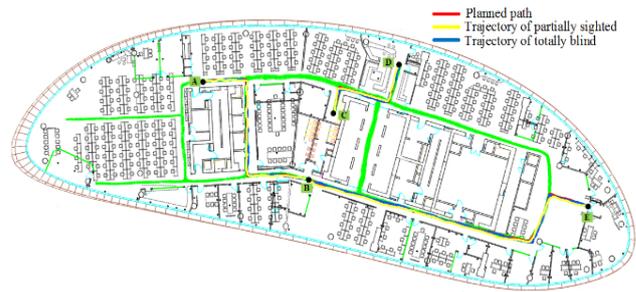

Fig. 8. Experiment 1 (without obstacles).

TABLE I
DISTANCE DEVIATING FROM THE PLANNED PATH (EXPERIMENT 1)

| Path | | Max. distance (m) | Average distance (m) | Variance(m$^2$) |
|---|---|---|---|---|
| A->B | Partially sighted | 0.2622 | 0.0967 | 0.0054 |
| | Totally blind | 0.3045 | 0.1523 | 0.0086 |
| D->C | Partially sighted | 0.3496 | 0.1124 | 0.0083 |
| | Totally blind | 0.4369 | 0.1938 | 0.0085 |
| E->B | Partially sighted | 0.6991 | 0.1152 | 0.0127 |
| | Totally blind | 0.6991 | 0.2249 | 0.0162 |

Experiment 2: this is to test the ability to solve both the route following and obstacle avoiding problems at the same time. The results shown in Fig. 9 verified that the effectiveness of the proposed navigation system. The user can avoid the obstacles and follow the globally shortest path to reach the destination. Similarly, in order to analyze the proposed algorithm quantitatively, the maximum and average deviation distances, as well as the variance were calculated, and shown in Table II. The maximum deviation distance of partially sighted is greater than that of the totally blind in AB and EB path, this is because the partially sighted individual has his own decision according to his observation, while the totally blind can only rely on the device and they usually do not deviate far from the path even when accounting obstacles. Differently, the partially sighted can keep going in straight line unless there are obstacles on his way, and the average deviation distance is less than the distance of the totally blind. The variance in Table II proved the above descriptions. This experiment verifies that the proposed navigation system can solve the route following and obstacle avoiding problems simultaneously.

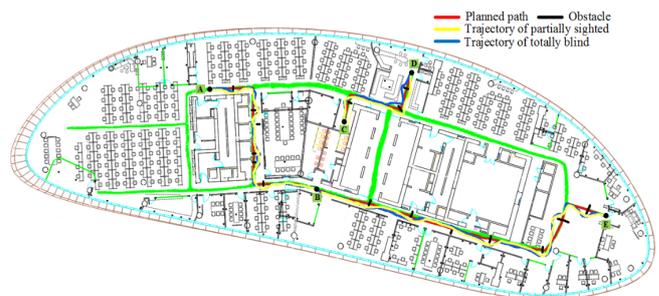

Fig. 9. Experiment 2 (with obstacles).

TABLE II
DISTANCE DEVIATING FROM THE PLANNED PATH (EXPERIMENT 2)

| Path | | Max. distance (m) | Average distance (m) | Variance(m$^2$) |
|---|---|---|---|---|
| A->B | Partially sighted | 1.1394 | 0.1553 | 0.0274 |
| | Totally blind | 0.9033 | 0.2920 | 0.0486 |
| D->C | Partially sighted | 0.7609 | 0.0984 | 0.0185 |
| | Totally blind | 1.3431 | 0.2960 | 0.0434 |
| E->B | Partially sighted | 1.0821 | 0.1692 | 0.0321 |
| | Totally blind | 1.0765 | 0.2897 | 0.0465 |

It is easy to see that the variance in the results of Experiment 2 is greater than that of Experiment 1. This is because the obstacles force the visually impaired individuals deviate from the path to prevent collisions. When they have avoided one obstacle, the proposed route following module will help them to get back to the pre-found the globally shortest virtual-blind-road. Both experiments verified the effectiveness of the proposed route following algorithm. Moreover, Experiment 2 proved that the proposed system could help the visually impaired individual reach the destination by following the globally shortest virtual-blind-road and meanwhile avoiding obstacles. The proposed navigation device proved to be able to serve as a consumer production for the ease of use in the visually impaired people's daily walks.

## VI. CONCLUSION

This paper presents a novel navigation device for the visually impaired groups to help them reach the destination safely and efficiently in indoor environment. The visual SLAM algorithm was used to solve the problems of indoor localization and virtual-blind-road building. The PoI-graph was generated to find a globally shortest virtual-blind-road by the A* based way finding algorithm. The dynamic sub-goal selecting based route following algorithm was proposed to help the blind follow the globally shortest virtual-blind-road as closely as possible and meanwhile avoid obstacles (including dynamic obstacles). Experimental results verified that the proposed navigation device was effective enough on helping the visually impaired people walk from one place to another. The sensors embedded on the device have the characteristics of low cost, small size and easy integration. Thus, it has great potential in consumer market, especially electronic travel aids market.

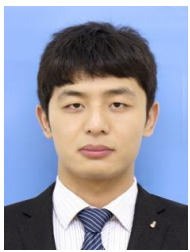

**Jinqiang Bai** got his B.E. degree and M.S. degree from China Uiversity of Petroleum in 2012 and 2015, respectively. He has been a Ph.D. student in Beihang University since 2015. His research interests include computer vision, deep learning, robotics, AI, etc.

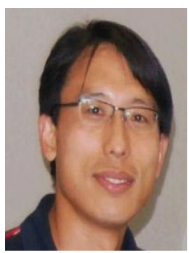

**Shiguo Lian** got his Ph.D. from Nanjing University of Science and Technology, China. He was a research assistant in City University of Hong Kong in 2004. From 2005 to 2010, he was a Research Scientist with France Telecom R&D Beijing. He was a Senior Research Scientist and Technical Director with Huawei Central Research Institute from 2010 to 2016. Since 2016, he has been a Senior Director with CloudMinds Technologies Inc. He is the author of more than 80 refereed international journal papers covering topics of artificial intelligence, multimedia communication, and human computer interface. He authored and co-edited more than 10 books, and held more than 50 patents. He is on the editor board of several refereed international journals.

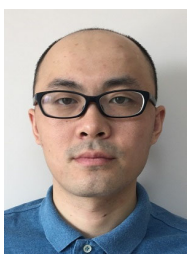

**Zhaoxiang Liu** received his B.S. degree and Ph.D. degree from the College of Information and Electrical Engineering, China Agricultural University in 2006 and 2011, respectively. He joined VIA Technologies, Inc. in 2011. From 2012 to 2016, he was a senior researcher in the Central Research Institute of Huawei Technologies, China. He has been a senior engineer in CloudMinds Technologies Inc. since 2016. His research interests include computer vision, deep learning, robotics, and human computer interaction and so on.

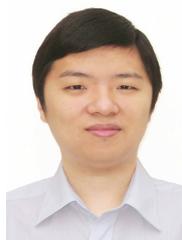

**Kai Wang** has been a senior engineer in CloudMinds Technologies Inc. since 2016. Prior to that, he was with the Huawei Central Research Institute. He received his Ph.D. degree from Nanyang Technological University, Singapore in 2013. His research interests include Augmented Reality, Computer Graphics, Human-Computer Interaction and so on. He has published more than ten papers on international journals and conferences.

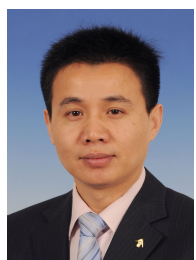

**Dijun Liu** has been the chief scientist of China academy of telecommunication technology (CATT), professor of Beihang University. He was the director of China institute of communications (CIC) and chairman of China communications integrated circuit committee (CCIC). He received his Ph.D. degree from China University of Petroleum (East China). He has 20+ years of prized academic research, industrial development, and entrepreneurship (chief scientist, Vice President, and CEO) in semiconductor and communication. He has won the 2016 National Science and technology progress special award by China State Council.